# Knowledge acquisition via interactive Distributed Cognitive skill Modules

Ahmet Orun

De Montfort University, Faculty of Computing, Engineering and media, Leicester UK, Email: aorun@dmu.ac.uk. Phone: +44(0)116 3664408

**Abstract**

The human's cognitive capacity for problem solving is always limited to his/her educational background, skills, experience, etc. Hence, it is often insufficient to bring solution to extraordinary problems especially when there is a time restriction. Nowadays this sort of personal cognitive limitation is overcome at some extent by the computational utilities (e.g. program packages, internet, etc.) where each one provides a specific background skill to the individual to solve a particular problem. Nevertheless these models are all based on already available conventional models or knowledge and unable to solve spontaneous unique problems, except human's procedural cognitive skills. But unfortunately such low-level skills can not be modelled and stored in a conventional way like classical models and knowledge. This work aims to introduce an early stage of a modular approach to procedural skill acquisition and storage via distributed cognitive skill modules which provide unique opportunity to extend the limits of its exploitation.

## 1 Introduction

The work presented here is strongly inspired by the ideas of "generic task" and "interaction problem" which were firstly introduced by Bylander and Chandrasekaran (1988). They indicate that the nature of problem and inference strategy strongly effect the knowledge acquisition and representation to solve a problem. The work also makes an emphasis on some kind of knowledge acquisition process which is capable of selecting particular knowledge for problem solving. They claim that this kind of knowledge acquisition may overcome the problems of expert systems. This work may be considered as an initial theoretic work which may support our proposal in terms of generic form of knowledge or cognitive skill acquisition via interaction.

In relation with the knowledge acquisition via proposed "novel" user-cognitive skill module interaction, we investigate the potential of procedural knowledge (cognitive skill) of each module user to solve the time constrained complicated problems such as "Minefield Navigation Task" introduced by Gordon et al. (1994). To bring a solution to such problems the work here aims to collect procedural type of knowledge via distributed interactive cognitive skill modules from unlimited number of users. The module would be design in same way of commercially available tools (e.g. Tetris, game console, Gameboy, etc.). The users should not be necessarily qualified or educated in a particular field to complete the task on the module. In further steps the knowledge collected from each module may be accumulated or integrated in a specially designed storage domain. This procedure will be like collecting and combining the cause-effect rules generated in each module to solve a certain problem or make a decision. Before we focus on how to realise a procedural knowledge (cognitive skill) acquisition and accumulation, we consider the most relevant works done on learning models of different knowledge types (e.g. procedural (intrinsic), declarative (expressible), etc.) and their relations with our proposed work. When module users develop certain skills as they interact with a specific task on the module, they also learn something while they uses their procedural knowledge to complete a specific cognitive task (e.g. game mission, math problem, etc.). Therefore there is close links between the learning models, declarative knowledge and cognitive skill developed during task completion.

Most conventional learning models are based on idea of turning declarative knowledge into procedural knowledge (cognitive skill) through practice. Sun et al. (2001) proposed bottom-up learning approach toward low-level skill gaining where the procedural knowledge develops before declarative knowledge. This indicates that in our proposed system the proposed cognitive module users should have no any pre-knowledge about the target task and may directly generate the procedural knowledge while they keep learning. This interaction will also lead to skill generation and collection via the module. Similarly the procedure of cognitive skill acquisition was pointed out by VanLehn (1995) which refers to learning by intellectually oriented tasks like puzzle solving, elementary geometry test, etc. Sun et al. (2001) indicated that the results obtained by these tasks would likely be transferable to real-world skill learning situations. In other example, the systems which use data warehouse (Adriaans et al.,1996) may also be in need of a cognitive skill to extract demanded knowledge in same way.

As we make an emphasis on efficiency of procedural knowledge, Anderson (1982, 1993) also stated that procedural knowledge was highly efficient property of human mind, once it has been developed it can work independently without need of declarative knowledge in many case. But so far less attention has been given to procedural knowledge assuming it is not externally accessible. Whereas our work proposes that procedural knowledge would be an alternative way to conventional communication between an individual and cognitive system domain (e.g. cognitive skill centre, personal cognitive device, etc.). Additionally the other advantages of procedural knowledge would be stated as follows :

1. It is more domain independent and be more easily adapted to any problem solving task.
2. More efficient to solve unique problems.
3. Since it is domain independent, it is more likely to be integrated with other counterparts from different domains and hence may be accumulated.
4. It does not require a priory knowledge during problem solving.

Several studies have been carried out on Procedural knowledge and learning. Reber and his co-workers (1967) investigated the acquisition of relatively complex rule systems by presenting their subjects a number of meaningless letter strings such as XVVCMS that were generated by artificial grammar. But with this experiments subjects were not informed about existence of a rule system. The results shoved that implicit learning is highly related to procedural knowledge. Since procedural knowledge has close link with unconscious mind, some authors pointed out its distinctive role. According to Kihlstrom (1987) unconscious mind is capable of abstracting rules, full semantic processing and even productive problem solving. All of these functions can take place outside of phenomenal awareness. Lewicki et al. (1992) stated that "our conscious thinking needs to rely on notes, computers, etc. to do the same job that our non-conscious operating algorithms can do instantly and without external help". Lamberts and Shanks (1997) also stated "This image of implicit non-conscious learning as a mechanism capable of unconscious rule abstraction, that may be even more powerful than our slow and capacity limited conscious strategies of knowledge acquisition".
These examples solely support our emphasis on the importance and potential of procedural knowledge and its possible role in skill acquisition and exploitation.

## 2 Materials and methods

To investigate the possibility of skill collection via user-cognitive module interaction, we consider the commercially available tools and similar utilities used to perform the specific tasks. One of them is Minefield Navigation Task (MNT) developed at the Naval Research Laboratory (1994). Here instead we use its modified version which is called COGNITE. In our task (Figure 1) an agent "A" needs to reach target "T" with its limited sensory information in a limited amount of time, by interacting with huge number (e.g. 100-200) of Unidentified Floating Objects (UFOs) in the field. The functions and behaviours of each UFO is totally unknown (whether it is hostile or alliance, how it is destroyed or exploited, what reactions it show in which case, etc.). Obviously a continuous journey for "A" to reach "T" is not possible in a certain time by the contribution of any human user or any software, overcoming such huge number of obstacles whose characteristics are yet unknown (unknown to any user or to any software). UFOs are represented by different shapes in Figure 1, and each one's secret characteristics are also listed in Table 1. The sensory utility for agent "A" provides shape and distance information of the objects.

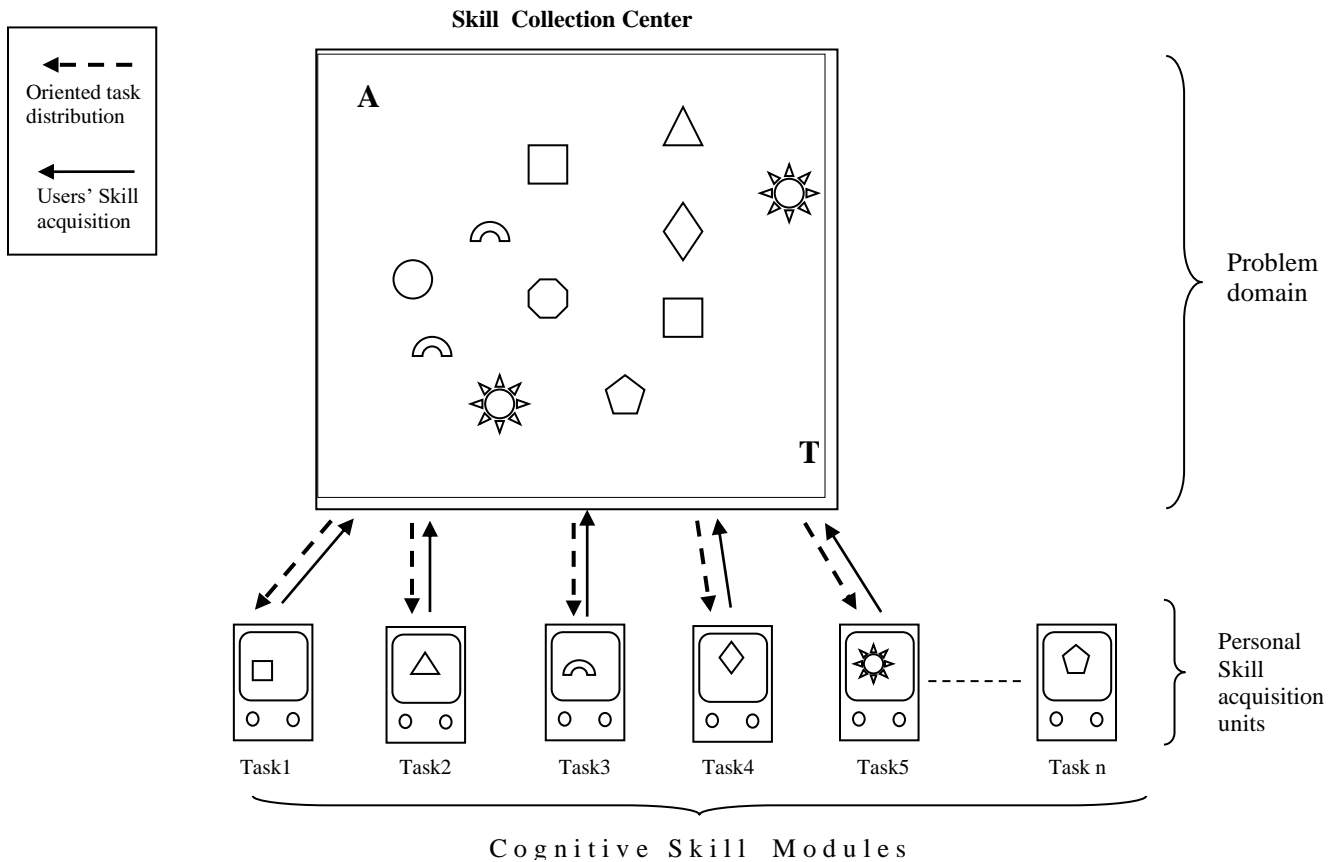

Figure 1. A "Mission impossible" task is sub divided and distributed over the skill acquisition units (cognitive modules) to collect and accumulate a personal cognitive skill which may be in cause-effect semantic (relational) format.

Each module user completes a unique task, reacting against a sequence of actions of UFOs in order to identify the characteristics and behaviours of them. The module user would be inexperience (preferably child) individual who has no previous training and he/she only follows the events in the module and reacts instantly while each reaction produces IF-THEN cause-effect rules which theoretically correspond to their cognitive skill maps. One module contains only one object whose characteristics are already unknown to database centre. The description of some objects and their characteristics are shown in Table 1.

Table 1. Some of the object types to dealt with by the module user and corresponding skill characteristics and attributions to be gained along the interaction process.

| Object type | Characteristics | Skill attributions gained by User-object Interaction |
|---|---|---|
| ▢ Destroyer | It destroys the agent if it is within certain range. But only effective along the horizontal and vertical axes. | To destroy it, approach along the directions of corners. Then fire at a certain distance. |
| ◯ Sticker | A sticky object. It attaches itself to the agent and diverts its control. | Avoid any contact. Push-and-pull to be release. |
| ☼ Power supply | It may be utilised to gain power. More power helps agent to increase its speed. | Simply keep touching it just for 2 seconds. |
| ⌒ Conveyor | It helps agent conveying it up to certain distance. | Touch any of two tips then it conveys you a few miles towards the target. |

As it was stated by Sun et al. (1996), cognitive skill is reactive sequential decision making. On the basis of this idea, in our work each module user takes prompt decision as he/she counteracts against the actions of the object. By these serial reactions, they learn and develop skill entirely independent from the main (central) task COGNITE whose complete inference strategy is located at database centre. Hence the users are unaware of their contribution to main task and the modules operate as interface between the users and database centre.

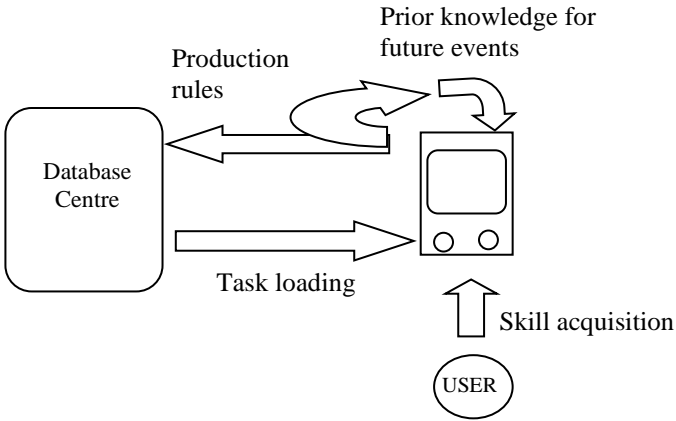

Figure 2. The cycle of knowledge acquisition and skill development. The module records the sequence of user's reactions against object's actions.

Since the user has no any pre knowledge about the task in module, he/she develops procedural knowledge and generates his/her own heuristics for the task. Like many other dynamic decision tasks (1992), the modules provide users with feedback on the outcomes of their actions. Hence, the users may use this outcome feedback to learn to improve their skill "on-line". The task learning curve converges by the time and users become more and more experienced (to gain full control over the object) and finally they constitute rules to complete the task. Generating these production rules from task instruction after repeated use is described as "proceduralization" which leads to skilled performance and studied by several authors (Anderson,1982; Rosenbloom et al. 1993). For each module user, analytical problem solving is common behaviour as it is related to three different aspects: classification, diagnosis and decision support (Lenz and Burkhard, 1998).

## 3 Results and discussion

The methodology of this specific application would be more generalised and applicable to other sorts of problem domain (e.g. trade, production, design, military, etc.) where a single or group of experts are unable to solve complicated problems within a time limit. This approach may also support available system domain such as expert systems to make them more stronger. The weakness of expert systems has been stated by Waterman (1986) and Negnevitsky (2002) in following reasons :

- Their data acquisition methods are not efficient and rather limited
- They are restricted to a very narrow domain of expertise.
- Since they are limited with in narrow domain, they are not flexible enough depending on certain rules.
- They need very efficient knowledge acquisition and storage
- It might take from five to ten person-years to build an expert system to solve a moderately difficult problem (for example MYCIN [18] took over 30 persons-years to build)

Hence, cognitive skill here may be promising to fill the gap between conventional expert systems and futuristic knowledge acquisition systems. But it should be noted that implementations suggested here are not restricted to expert systems. They would lead to better analysis of cognitive capacity of human brain, new approach to database systems, etc.

Within this work the suggestion of cognitive skill module (hardware & software) is capable of demonstrating a special task, which encourages user to produce cognitive skill. An ideal integration of everybody's cognitive skill is only be possible via social "brain network" between the individual as it may already be demonstrated in social domains in daily life but unable to control and store its cognitive outputs. The proposed system "cognitive skill module" may help reach such similar target at some extent.

As a conclusion we make the following remarks about the cognitive skill

- Unique cognitive skill is a skill which is developed uniquely by users' unique reactions against unique stimuli event in the module domain (such unique information may not be available in any domain on the Earth)
- The best way of making use of cognitive skill is collecting it from different individuals by specially designed interactive tools like personal device.
- All cognitive skill modules have equal weights (priorities) disregarding their domain conditions or each individual's qualification.

# 4 Possibility of future works and further evaluations

Within this article our suggestion is to focus on a concept called Cognitive Skill, which spontaneously reveals in every day's action of our brain without our awareness. But they are not adequately exploited and stored since they are not collected nor merged with each other in a conventional format but rather remain at non-modelled deepest level. Here the suggestion has been made that if such skill material would be stored and exploited, we would have a unique opportunity to build up futuristic systems. Even perhaps in near future this kind of cognitive skill will be sold (like electricity) at the collection centres in a form of millions of "cause-effect rules" at a specific bandwidth ready to flow into any customer domain. And perhaps in near future for any high-level problem solution we will just get connected to our personal "cognitive skill socket" in our office. Or the idea of cognitive skill modules will open a new era of self-employment, where any individual will be able to sell his/her 5-10 minute brain power to the "skill centres" via specially designed outdoor collection points.